\documentclass[11pt, a4paper]{article}
\usepackage{emnlp2017}

\usepackage[english]{babel}

\usepackage{amsmath}
\usepackage{graphicx}
\usepackage{pslatex}
\usepackage{latexsym}
\usepackage{xspace}
\usepackage{multirow}
\usepackage{amsfonts}
\usepackage{caption}
\usepackage{tabularx}
\usepackage{pbox}
\usepackage{color}
\usepackage{xcolor}
\usepackage{url}
\usepackage{xargs} 
\usepackage[colorinlistoftodos,prependcaption,textsize=scriptsize]{todonotes}

\emnlpfinalcopy

\def\emnlppaperid{644}

\newcommand{\tabref}[2][]{Table#1~\ref{#2}\xspace}

\newcommand{\para}{\textsc{Parallel}\xspace}
\newcommand{\pivot}{\textsc{Pivot}\xspace}
\newcommand{\parasym}{\textsc{Parallel-Sym}\xspace}
\newcommand{\pivotsym}{\textsc{Pivot-Sym}\xspace}
\newcommand{\paraasym}{\textsc{Parallel-Asym}\xspace}
\newcommand{\pivotasym}{\textsc{Pivot-Asym}\xspace}

\title{Image Pivoting for Learning Multilingual Multimodal Representations}

\author{
Spandana Gella \quad Rico Sennrich \quad Frank Keller \quad Mirella Lapata\\
Institute for Language, Cognition and Computation \\
School of Informatics, University of Edinburgh \\
\{spandana.gella, rico.sennrich\}@ed.ac.uk \\ 
\{keller,mlap\}@inf.ed.ac.uk
}

\begin{document}
\maketitle

\begin{abstract}
In this paper we propose a model to learn multimodal 
multilingual representations for matching 
images and sentences in different languages, 
with the aim of advancing multilingual 
versions of image search and image understanding.
Our model learns a common representation for
images and their descriptions in two different languages (which need not be parallel) by considering the image 
as a pivot between two languages.
We introduce a new pairwise ranking loss function 
which can handle both symmetric and asymmetric 
similarity between the two modalities.
We evaluate our models on image-description ranking for 
German and English, and on semantic textual similarity of image descriptions in English. 
In both cases we achieve state-of-the-art performance.

\end{abstract}

\section{Introduction}

In recent years there has been a significant amount of 
research in language and vision tasks which require the 
joint modeling of texts and images. Examples include text-based 
image retrieval, image description and visual question answering. 
An increasing number of large
image description datasets has become available
\cite{hodosh:flickr8k:2013,young:flickr30k:2014,mscoco:2014}
and various systems have been proposed to handle the image description 
task as a generation problem \cite{Bernardi:ea:16,mao2014deep,google:show-tell:2014,fang:msr:captions:2015}.
There has also been a great deal of work on sentence-based image 
search or cross-modal 
retrieval where the objective is to learn a joint 
space for images and text 
\cite{hodosh:flickr8k:2013,devise:2013,karpathy:deep-fragments:2014,kiros2014unifying,sdt-rnn:2015,lrcn:2015}. 

Previous work on image description generation 
or learning a joint space for images 
and text has mostly focused on English due to 
the availability of English datasets. 
Recently there have been attempts to create image descriptions and 
models for other languages \cite{uiuc:jp:2015,multi30k:2016,bridgeCorr:2016,yjcaptions:2016,specia:mmt:2016,li:flickr8kcn:2016,imagePivotACL16,YoshikawaST17}. 

Most work on learning a joint space for images and their 
descriptions is based on Canonical Correlation Analysis (CCA) 
or neural variants of CCA over representations of image and its descriptions \cite{hodosh:flickr8k:2013,andrew:dcca:2013,yan:dcca:2015,gong:tcca:2014, chandar:corrnet:2016}. 
Besides CCA, a few others learn a visual-semantic or 
multimodal embedding space of image descriptions and  
representations by optimizing a ranking cost function 
\cite{kiros2014unifying,sdt-rnn:2015,ma2015multimodal,vendrov2015order} 
or by aligning image regions (objects) 
and segments of the description 
\cite{karpathy:deep-fragments:2014,plummer2015flickr30k} 
in a common space. Recently \citet{lin:leveraging:vqa:embeddings:2016} 
have leveraged visual question answering models to encode images and
descriptions into the same space.

However, all of this work is targeted at monolingual 
descriptions, i.e., mapping images and descriptions 
in a single language onto a joint embedding space.
The idea of pivoting or bridging is not new and language 
pivoting is well explored for machine 
translation \cite{wu2007pivot,FiratSAYC16} and to learn multilingual multimodal representations \cite{bridgeCorr:2016,calixto:mlmme:2017}.
\citet{bridgeCorr:2016} propose a model to learn 
common representations between $M$ views and assume 
there is parallel data available between a pivot view 
and the remaining $M-1$ views. Their multimodal experiments 
are based on English as the pivot and 
use large parallel corpora available between 
languages to learn their representations.

Related to our work \citet{calixto:mlmme:2017} proposed a model 
for creating multilingual multimodal embeddings. 
Our work is different from theirs in that we choose the
image as the pivot and use a different similarity function. 
We also propose a single model for 
learning representations of images and multiple languages,
whereas their model is language-specific.

  In this paper, we
learn multimodal representations in multiple languages, i.e., our
model yields a joint space for images and text in multiple languages
using the image as a pivot between languages. 
We propose a new objective function in a multitask 
learning setting and jointly optimize the mappings 
between images and text in two different languages. 

\begin{figure}[t]
\centering
\includegraphics[trim={0 2 0 0},clip, height=36mm,width=\columnwidth]{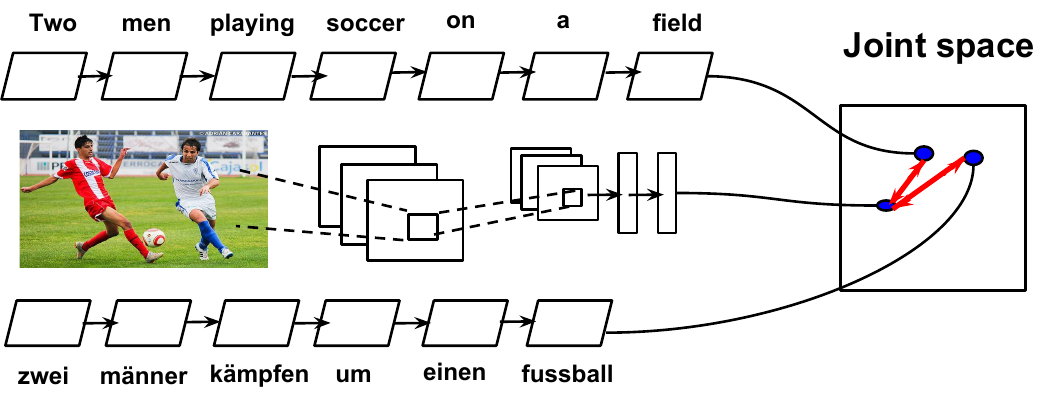}
\caption{Our multilingual multimodal model with image as pivot}
\label{fig:proposed_model}
\end{figure}

\section{Dataset}
\label{sec:dataset}
We experiment with the Multi30k dataset, a multilingual 
extension of Flickr30k corpus \cite{young:flickr30k:2014} consisting of 
English and German image descriptions \cite{multi30k:2016}.
The Multi30K dataset has 29k, 1k and 1k images in the train, validation and test splits respectively,
and contains two types of multilingual annotations: 
(i) a corpus of one English description per image and its translation 
into German; and (ii) a corpus 
of five independently collected English and German descriptions 
per image. We use the independently collected English and 
German descriptions to train our models. Note that these 
descriptions are not translations of each other, i.e., 
they are not parallel, although they describe the same image.

\section{Problem Formulation}
Given an image $i$ and its descriptions $c_1$ and $c_2$ in 
two different languages our aim is to learn a model 
which maps $i$, $c_1$ and $c_2$
onto same common space $\mathbb{R}^{N}$ 
(where $N$ is the dimensionality of the embedding space) 
such that the image and its gold-standard descriptions in 
both languages are mapped close to each other 
(as shown in Figure~\ref{fig:proposed_model}). 
Our model consists of the embedding functions 
$f_i$ and $f_{c}$ to encode images and descriptions and
a scoring function $S$ to compute the similarity 
between a description--image pair. 

In the following we describe two models: (i)~the \pivot model that uses the image 
as pivot between the description in both the languages; (ii)~the \para model 
that further forces the image descriptions in both languages to be closer to each other in the joint space.
We build two variants of \pivot and \para with different similarity functions $S$ to learn the joint space.

\subsection{Multilingual Multimodal Representation Models}
\label{sec:model}
In both \pivot  and \para  we use a deep  
convolutional neural network architecture (CNN) 
to represent the image $i$ 
denoted by $f_i(i)= W_{i} \cdot CNN(i)$ where $W_{i}$ is 
a learned weight matrix and $CNN(i)$ 
is the image vector representation. For each language 
we define a recurrent neural network 
encoder $f_c(c_{k}) = GRU(c_{k})$ 
with gated recurrent units (GRU) activations to encode the description $c_{k}$.

In \pivot, we use 
monolingual corpora from multiple languages of sentences aligned with images to learn the joint space.
The intuition of this model is that an image is a universal representation across all languages, and if we constrain a sentence representation to be closer to image, sentences in different languages may also come closer.
Accordingly we design a loss function as follows:
\begin{equation} 
\small
\begin{split}
loss_{pivot} = \sum_{k} \bigg[ \sum_{(c_{k}, i)} \bigg( \sum_{c'_{k}} \max\{0, \alpha - S(c_{k},i) + S(c'_{k}, i)\}  \\
   + \sum_{i'} \max\{0, \alpha - S(c_{k},i) + S(c_{k}, i')\} \bigg) \bigg]
\end{split}
\label{eq:contrastive} 
\end{equation}
\noindent where $k$ stands for each language.
This loss function encourages the similarity $S(c_{k},i)$ between gold-standard description $c_{k}$ and image $i$ to be greater than any other irrelevant description $c'_{k}$ by a margin $\alpha$.
A similar loss function is useful for learning 
multimodal embeddings in a single language \cite{kiros2014unifying}.
For each minibatch, we obtain invalid descriptions by selecting descriptions of other images except the current image of interest and vice-versa.

In \para, in addition to making an image similar to a description, we make multiple descriptions of the same image in different languages similar to each other, based on the assumption that these descriptions, although not parallel, share some commonalities.
Accordingly we enhance the previous loss function with an additional term:
\begin{equation} 
\small
\begin{split}
loss_{para} = loss_{pivot} + 
\sum_{(c_{1}, c_{2})} \bigg( \sum_{c'_{1}} \max\{0, \alpha - S(c_{1},c_{2}) \\
\noindent + S(c'_{1}, c_{2})\} + \sum_{c'_{2}}\max\{0, \alpha - S(c_{1},c_{2}) + S(c_{1}, c'_{2})\} \bigg) 
\end{split}
\label{eq:contrastive-lang} 
\end{equation} 
\noindent Note that we are iterating over all pairs of descriptions $(c_1, c_2)$, and maximizing the similarity between descriptions of the same image and at the same time minimizing the similarity between descriptions of different images.

We learn models using two similarity functions:  symmetric and asymmetric. 
For the former we use cosine similarity
and for the latter we use the metric of \newcite{vendrov2015order}
which is useful for learning embeddings that maintain  an order, e.g., dog and cat are more closer to pet than animal while being distinct.
Such ordering is shown to be useful in building effective multimodal space of images and texts.
An analogy in our setting would be two descriptions of an image are closer to the image while at the same time preserving the identity of each (which is useful when sentences describe two different aspects of the image).
The similarity metric is defined as:
\begin{equation} \small 
S(a,b) = - ||max(0, b - a)||^2 \label{eq:order-score}
\end{equation}
\noindent where $a$ and $b$ are embeddings of image and description.

We call the symmetric similarity variants of our models as \pivotsym and \parasym, and the asymmetric variants \pivotasym and \paraasym.

\section{Experiments and Results}

We test our model on the tasks of image-description ranking and
semantic textual similarity.  We work with each language separately.
Since we learn embeddings for images and languages in the same
semantic space, our hope is that the training data for each modality
or language acts complementary data for the another modality or
language, and thus helps us learn better embeddings.

\paragraph{Experiment Setup}

We sampled minibatches of size 64 images and their descriptions, and drew all negative samples from the minibatch.
We trained using the Adam optimizer with learning 
rate 0.001, and early stopping on the validation set.
Following \citet{vendrov2015order} we set the dimensionality of the embedding space and the GRU hidden layer $N$ to 1024 for both English and German. 
We set the dimensionality of the learned word embeddings to 300 for both languages, and 
the margin $\alpha$ to 0.05 and 0.2, respectively, to learn asymmetric and symmetric similarity-based embeddings.\footnote{We constrain the embeddings of descriptions and images to have non-negative entries when using asymmetric similarity by taking their absolute value.} We keep all hyperparameters constant across all models.
We used the L2 norm to mitigate over-fitting \cite{kiros2014unifying}. 
We tokenize and truecase both English and German  descriptions using the Moses Decoder scripts.\footnote{\url{https://github.com/moses-smt/mosesdecoder/tree/master/scripts}}

\begin{table*}[t]
\begin{minipage}{\columnwidth}
\begin{center}
\resizebox{\columnwidth}{!}{
\begin{tabular}{l@{~}r@{~}r@{~}r@{~}r@{~}rr@{~}r@{~}r@{~}r@{~}}
\hline
 System&  \multicolumn{4}{c}{Text to Image} & &\multicolumn{4}{c}{Image to Text} \\
\cline{2-5} \cline{7-10}
&  R@1 & R@5 & R@10 &Mr & & R@1 & R@5 & R@10 &Mr \\
\cline{2-5} \cline{7-10}
VSE \small{\cite{kiros2014unifying}}  & 23.3 & 53.6 & 65.8 &5 && 31.6 & 60.4 & 72.7  &3\\ 
OE \small{\cite{vendrov2015order}}& 25.8 & \textbf{56.5} &67.8 &4  && \textbf{34.8} & \textbf{63.7} & 74.8 &3 \\ 
\hline
\pivotsym   & 23.5 & 53.4 & 65.8 &5 && 31.6 & 61.2 & 73.8 &3\\
\parasym   & 24.7 & 53.9 & 65.7 &5 && 31.7 & 62.4 & 74.1 &3\\ 
\pivotasym  & 26.2 & 56.4 & \textbf{68.4} &4 && 33.8 & 62.8 & \textbf{75.2} &3 \\ 
\paraasym  & \textbf{27.1} & 56.2 & 66.9&4  && 31.5  & 61.4 & 74.7 &3 \\ %
 \hline
\end{tabular}
}
\caption{Image-description ranking results of \mbox{English} on Flickr30k test data.}
\label{tab:english-embedding-results}
\end{center}
\end{minipage}
\hspace{1em}
\begin{minipage}{\columnwidth}
\resizebox{\columnwidth}{!}{
\begin{tabular}{l@{~}r@{~}r@{~}r@{~}r@{~}lr@{~}r@{~}r@{~}r@{~}}
\hline
 System& \multicolumn{4}{c}{Text to Image} & & \multicolumn{4}{c}{Image to Text} \\
\cline{2-5} \cline{7-10}
&  R@1 & R@5 & R@10 &Mr  && R@1 & R@5 & R@10  & Mr \\
\cline{2-5} \cline{7-10}
VSE \small{\cite{kiros2014unifying}} &  20.3 & 47.2 & 60.1 &6 && 29.3 & 58.1 & 71.8 &4\\ 
OE \small{\cite{vendrov2015order}} &  21.0 &48.5 &60.4 &6  && 26.8 &57.5 & 70.9 &4\\ 
\hline
\pivotsym  &  20.3 & 46.4 & 59.2 &6&&  26.9 & 56.6 & 70.0 &4\\ 
\parasym & 20.9 & 46.9 &59.3 &6 && 28.2 & 57.7 & 71.3 &4 \\
\pivotasym & \textbf{22.5} & 49.3 & 61.7 &6 && 28.2 & \textbf{61.9} & \textbf{73.4} & \textbf{3}\\  
\paraasym & 21.8 & \textbf{50.5} & \textbf{62.3} & \textbf{5} && \textbf{30.2} & 60.4 & 72.8 & \textbf{3}\\ 
\hline  
\end{tabular}
}
\caption{Image-description ranking results of \mbox{German} on Flickr30k test data. }
\label{tab:german-embedding-results}
\end{minipage}
\end{table*}

\begin{table*}
\resizebox{2.0\columnwidth}{!}{
\tabcolsep 1.5pt
\begin{tabular}{lp{10cm}crrr}
\hline
\small{Image} & \small{Descriptions} &  & \multicolumn{3}{c}{\small{Image Rank}}  \\
\hline
& & & \scriptsize{OE} & \scriptsize{\pivot} & \scriptsize{\para} \\
\cline{4-6} 
\multirow{ 2}{*}{\includegraphics[trim={0 2 0 0},clip, height=17mm,width=0.35\columnwidth]{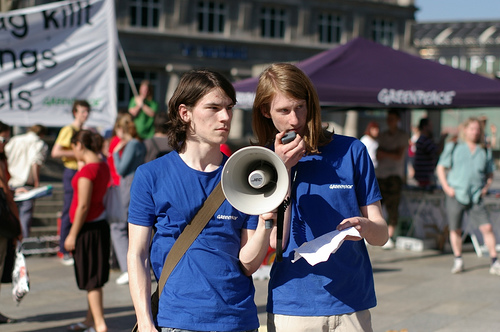}} & \small{2 Menschen auf der Stra{\ss}e mit Megafon} \vspace{0.3cm} & & \small{141} & \small{37} & \small{6} \\
 & \small{two people in blue shirts are outside with a bullhorn} \vspace{0.3cm} & & \small{85} & \small{7} & \small{3}  \\
\\
\multirow{ 2}{*}{\includegraphics[trim={0 2 0 0},clip, height=17mm,width=0.35\columnwidth]{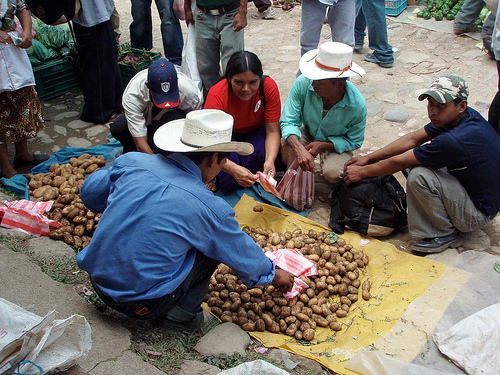}} 
& \small{ein Verk\"{a}ufer mit wei{\ss}em Hut und blauem Hemd , verkauft Kartoffeln oder \"{a}hnliches an M\"{a}nner und Frauen} \vspace{0.1cm}
& & \small{36} & \small{1} & \small{3} \\
& \small{at an outdoor market , a small group of people stoop to buy potatoes from a street vendor , who has his goods laid out on the ground} \vspace{0.1cm} & & \small{24} & \small{2} & \small{2} \\
  \hline
\end{tabular}
}
\caption{The rank of the gold-standard image when using each German and English descriptions as a query on models trained using asymmetric similarity.}
\label{tab:model-analysis}
\end{table*}

To extract image features, we used a convolutional neural network model trained on 1.2M images of 1000 class ILSVRC~2012 object classification dataset, a subset of ImageNet \cite{imagenet:2014}. 
Specifically, we used VGG 19-layer CNN architecture and
extracted the activations of the penultimate fully 
connected layer to obtain features for all images in the 
dataset \cite{vgg:2014}. We use average features  from 10 crops of the
re-scaled images.\footnote{We rescale images so that the smallest side
  is 256 pixels wide, we take 224 $\times$ 224 crops from the 
corners, center, and their horizontal reflections to get 
10 crops for the image.}

\paragraph{Baselines}
As baselines we use monolingual models, i.e., models trained 
on each language separately.
Specifically, we use Visual Semantic Embeddings (VSE) 
of \newcite{kiros2014unifying} and Order Embeddings (OE) of \newcite{vendrov2015order}. 
We use a publicly available implementation to train both VSE and OE.\footnote{\url{https://github.com/ivendrov/order-embedding}}

\subsection{Image-Description Ranking Results} 
To evaluate the multimodal multilingual embeddings, we report 
results on an image-description ranking task. Given a query 
in the form of a description or an image, the task its to retrieve all images or 
descriptions sorted based on the relevance. 
We use 
the standard ranking evaluation metrics 
of recall at position $k$ (R@K, where higher is better) and 
median rank (Mr, where lower is better)
to evaluate our models. 
We report results for both English
and German descriptions. Note that we have one single model for both languages.

In Tables~\ref{tab:english-embedding-results} and~\ref{tab:german-embedding-results} 
we present the ranking results of the baseline models of \citet{kiros2014unifying} and 
\citet{vendrov2015order} and our proposed \pivot and \para models. We do not compare our image-description ranking 
results with \citet{calixto:mlmme:2017} since they report results on half of validation set of Multi30k whereas our results are on the publicly available test set of Multi30k.
For English, \pivot with asymmetric similarity is either  competitive or better than monolingual models and symmetric similarity, especially in the R@10 category it obtains state-of-the-art.  
For German, both \pivot and \para with the 
asymmetric scoring function outperform monolingual models and symmetric similarity. 
We also observe that the German ranking experiments benefit the most from the multilingual
signal. A reason for this could be that the German description corpus
has many singleton words (more than $50\%$ of the vocabulary) and
English description mapping might have helped in learning better
semantic embeddings.  These results suggest that the multilingual signal
could be used to learn better multimodal embeddings, irrespective of the
language.  
Our results also show that the asymmetric scoring function can help learn better embeddings. In \tabref{tab:model-analysis} we present a few examples where 
\pivotasym and \paraasym models performed better on both the languages 
compared to baseline order embedding model even using descriptions of very different lengths as queries.

\begin{table}[t]
\resizebox{\columnwidth}{!}{
\begin{tabular}{lrc@{~}c@{~}c@{~}}
\hline
Model & VF & 2012 & 2014 & 2015 \\
\hline
Shared Task Baseline & $-$& 29.9&$51.3$ & $60.4$\\
STS Best System & $-$ & 87.3&$83.4$ & $86.4$ \\
GRAN \cite{wieting2017learning} & $-$ & 83.7 & 84.5 & 85.0\\
\hline
MLMME \cite{calixto:mlmme:2017} & VGG19 & $-$&$72.7$ & $79.7$ \\
VSE \cite{kiros2014unifying} & VGG19 & 80.6&82.7 & 89.6 \\
OE \cite{vendrov2015order} & VGG19 & 82.2& 84.1 & 90.8 \\
\hline
\pivotsym & VGG19 & 80.5&81.8 & 89.2\\
\parasym & VGG19&  82.0&81.4 & 90.4\\
\pivotasym & VGG19 & 83.1& 83.8 & 90.3  \\
\paraasym & VGG19& \textbf{84.6} & \textbf{84.5} & \textbf{91.5}\\
\hline
\end{tabular}
}
\caption{Results on Semantic Textual Similarity Image datasets (Pearson's $r$ $\times$ 100 ). Our systems that performed better than best reported shared task scores are in \textbf{bold}.}
\label{tab:sts-results}
\end{table}

\subsection{Semantic Textual Similarity Results}
In the semantic
textual similarity task (STS), we use the textual embeddings from
our model to compute the similarity between a pair of 
sentences (image descriptions in this case). We evaluate on video task from STS-2012 and image tasks from STS-2014, STS-2015
(\citealt{sts:2012}, \citealt{sts:2014}, \citealt{sts:2015}). 
The video descriptions in the STS-2012 task are 
from the MSR video description corpus \cite{chen:msrvid:2011} and the image 
descriptions in STS-2014 and 2015 are from UIUC PASCAL 
dataset \cite{uiuc:pascal:2010}.

In~\tabref{tab:sts-results}, we present the Pearson correlation 
coefficients of our model predicted scores with the gold-standard 
similarity scores provided as part of the STS image/video description tasks.
We compare with the best reported scores for the 
STS shared tasks, achieved by MLMME \cite{calixto:mlmme:2017}, paraphrastic sentence embeddings \cite{wieting2017learning}, 
visual semantic embeddings 
\citep{kiros2014unifying}, and order 
embeddings \citep{vendrov2015order}. 
The shared task baseline is computed based on word overlap 
and is high for both the 2014 and the 2015 dataset, indicating that 
there is substantial lexical overlap between the STS image description datasets. 
Our models outperform both the baseline 
system and the best system submitted to the 
shared task. For the 2012 video paraphrase corpus, our 
multilingual methods performed better than the monolingual 
methods showing that similarity across paraphrases can be 
learned using multilingual signals. Similarly, \citet{wieting2017learning} have reported to learn better paraphrastic sentence embeddings with multilingual signals.
Overall, we observe that models learned using 
the asymmetric scoring function outperform the state-of-the-art on
these datasets, suggesting that multilingual sharing is beneficial.
Although the task has nothing to do German, because our models can make use of datasets from different languages, we were able to train on significantly larger training dataset of approximately~145k descriptions.
\citet{calixto:mlmme:2017} also train on a larger dataset like ours,
but could not exploit this to their advantage. In
\tabref{tab:sts-analysis} we present the 
example sentences with the highest and lowest difference between gold-standard and predicted semantic textual similarity scores using our best performing \paraasym model.

\begin{table}[t]
\tabcolsep 2pt
\resizebox{\columnwidth}{!}{
\begin{tabular}{p{3cm}p{3cm}cc}
\hline
\small{S1} & \small{S2} & \small{GT} & \small{Pred} \\
\hline
\small{Black bird standing on concrete.} & \small{Blue bird standing on green grass.} & \small{1.0} & \small{4.2}\\
\small{Two zebras are playing.} & \small{Zebras are socializing.} & \small{4.2} & \small{1.2}\\
\hline
\small{Three goats are being rounded up by a dog.} & \small{Three goats are chased by a dog} & \small{4.6} & \small{4.5} \\
\small{A man is folding paper.} & \small{A woman is slicing a pepper.} & \small{0.6} & \small{0.6}\\
\hline
\end{tabular}
}
\caption{Example sentences with gold-standard semantic textual similarity score and the predicted score using our best performing \paraasym model.}
\label{tab:sts-analysis}
\end{table}

\section{Conclusions}
We proposed a new model that jointly learns multilingual multimodal representations using the image as a pivot between languages.
We introduced new objective functions that can exploit similarities between images and descriptions across languages. 
We obtained state-of-the-art results on two tasks: image-description ranking and semantic textual similarity.
Our results suggest that exploiting multilingual and multimodal resources can help in learning better semantic representations.

\section*{Acknowledgments}
This work greatly benefited from discussions
with Siva Reddy and Desmond Elliot. The authors would like to 
thank the anonymous reviewers for their helpful comments. 
The authors gratefully acknowledge the support of the 
European Research Council (Lapata: award number 681760).

\bibliographystyle{emnlp_natbib}
\bibliography{references}

\end{document}